\documentclass[conference]{IEEEtran}
\IEEEoverridecommandlockouts
\usepackage{cite}
\usepackage{hyperref}
\usepackage{amsmath,amssymb,amsfonts}
\usepackage{algorithmic}
\usepackage{graphicx}
\usepackage{textcomp}
\usepackage{xcolor}
\usepackage[ruled]{algorithm2e}
\usepackage[labelformat=simple]{subcaption}
\def\BibTeX{{\rm B\kern-.05em{\sc i\kern-.025em b}\kern-.08em
    T\kern-.1667em\lower.7ex\hbox{E}\kern-.125emX}}

\usepackage{fancyhdr}
\usepackage{lipsum} 

\fancypagestyle{plain}{
  \fancyhf{}
  \fancyfoot[C]{\thepage}%
}
\fancypagestyle{firstpage}{
  \fancyhf{}
  \fancyhead[L]{\small Accepted version of the paper presented at 2022 Australasian Universities Power Engineering Conference (AUPEC), Adelaide, Australia}
  \fancyfoot[L]{\small © 2022 IEEE.  Personal use of this material is permitted.  Permission from IEEE must be obtained for all other uses, in any current or future media, including reprinting/republishing this material for advertising or promotional purposes, creating new collective works, for resale or redistribution to servers or lists, or reuse of any copyrighted component of this work in other works.}%
}
	\usepackage{etoolbox}
	\makeatother
	\makeatletter
	\patchcmd{\@maketitle}
	 {\addvspace{0.5\baselineskip}\egroup}
	 {\addvspace{-1\baselineskip}\egroup}
	 {}
	 {}
\makeatother

\begin{document}

\title{Efficient anomaly detection method for rooftop PV systems using big data and permutation entropy}
\author{\\[-4ex]
\IEEEauthorblockN{Sahand Karimi-Arpanahi}
\IEEEauthorblockA{\textit{School of Electrical and Electronic Engineering} \\
\textit{The University of Adelaide} \\
Adelaide, Australia\\
\href{mailto:sahand.karimi-arpanahi@adelaide.edu.au}{sahand.karimi-arpanahi@adelaide.edu.au} }
\and\\[-4ex]
\IEEEauthorblockN{S. Ali Pourmousavi}
\IEEEauthorblockA{\textit{School of Electrical and Electronic Engineering} \\
\textit{The University of Adelaide} \\
Adelaide, Australia\\
\href{mailto:a.pourm@adelaide.edu.au}{a.pourm@adelaide.edu.au}}
}
\maketitle
\thispagestyle{firstpage}

\begin{abstract}
The number of rooftop photovoltaic (PV) systems has significantly increased in recent years around the globe, including in Australia. This trend is anticipated to continue in the next few years. Given their high share of generation in power systems, detecting malfunctions and abnormalities in rooftop PV systems is essential for ensuring their high efficiency and safety. In this paper, we present a novel anomaly detection method for a large number of rooftop PV systems installed in a region using big data and a time series complexity measure called weighted permutation entropy (WPE). This efficient method only uses the historical PV generation data in a given region to identify anomalous PV systems and requires no new sensor or smart device. Using a real-world PV generation dataset, we discuss how the hyperparameters of WPE should be tuned for the purpose. The proposed PV anomaly detection method is then tested on rooftop PV generation data from over 100 South Australian households. The results demonstrate that anomalous systems detected by our method have indeed encountered problems and require a close inspection. The detection and resolution of potential faults would result in better rooftop PV systems, longer lifetimes, and higher returns on investment.
\end{abstract}


\section{Introduction}
Global climate change has urged the transition from fossil fuel generation to a low-carbon electricity grid supplied by renewable energy resources. To this end, countries have adopted various approaches toward a greener power grid. In Australia, solar photovoltaic (PV) systems have become an important part of clean electricity generation, producing more than 10\% of electricity in 2021 and being the fastest growing generation type in Australia \cite{CEC2022, arenasolar}. Currently, at least every 1 out of 4 houses in Australia and almost 50\% of the residential consumers in South Australia (SA) are equipped with rooftop PV systems \cite{Solarfeeds}. In 2020 and 2021, Australian homes and businesses installed record amounts of rooftop PV systems, adding around 3 GW of capacity each year, even in the face of the pandemic \cite{REcon}. With the ambitious renewable energy targets set by state and federal governments, this rapid integration of rooftop PV systems is set to continue in the coming years\cite{GovProg}. 

Despite the positive aspects of higher rooftop PV system penetration in the power grid, monitoring their performance individually is challenging for PV system service providers and their operation and maintenance (O\&M) crews. This is because rooftop PV systems are widely dispersed, and they do not possess costly sensors and measurement devices that are necessary to accurately track their performance. Nonetheless, these systems are prone to many types of partially or fully malfunctioning, e.g., faulty PV panel or electrical connections, partial/seasonal shading, and faulty inverters  \cite{pillai2018comprehensive}. These issues may lead to significant operational issues for the system operators and a significant drop in efficiency, reliability and safety of them for the owners \cite{zhao2013outlier}. Therefore, it is imperative to ensure the faultless operation of PV systems.

Numerous studies have been conducted over the past few years to propose efficient and reliable detection methods for PV system faults. In \cite{firth2010simple}, authors developed an empirical method for finding PV system faults, using the relationship between solar irradiance and the PV system performance. Authors in \cite{kumar2017online} proposed a method for detecting one type of electrical faults in PV arrays based on the wavelet transforms of the PV array voltage and current, which are computationally intensive solutions. In \cite{harrou2018reliable}, Harrou \textit{et al.} developed a statistical method for detecting and diagnosing some PV system faults based on a single-diode model, using its voltage, current, and measured temperature and irradiance. Using similar inputs, a machine-learning-based method was proposed in \cite{madeti2018modeling} for PV fault detection and diagnosis. Authors in \cite{khoshnami2018two} presented a fault detection algorithm for the DC side of PV systems by using the array voltage and current, which required an intelligent electronic device for each PV system to process the required complicated calculations. In \cite{fadhel2019pv}, Fadhel \textit{et al.} developed a detection and classification method for PV shading, which used the PV array voltage, current, temperature, and irradiance. Authors in \cite{pillai2018mppt} proposed a ``sensorless'' detection method for line-to-line and line-to-ground faults in PV arrays, which only required their voltage and current measurements. In \cite{iqbal2021real}, a fault detection method was developed for large-scale grid-tied PV farms, which requires irradiance, temperature and generated power as the input. Yurtseven \textit{et al.} presented a sensorless method in \cite{yurtseven2021sensorless} for fault detection in solar PV farms, using the voltage and current of all inverters in a PV farm. 

The above papers are just a handful of the existing studies on PV system fault detection. As discussed comprehensively in \cite{pillai2018comprehensive}, these studies can be classified into several groups, each with specific advantages and drawbacks. For example, some methods, e.g., \cite{firth2010simple, harrou2018reliable, madeti2018modeling, fadhel2019pv, iqbal2021real}, require new sensors for measuring the temperature and irradiance. This is not only an expensive solution for a residential rooftop PV system but also needs access to all PV systems dispersed around the country. The other group of studies, which proposed sensorless fault detection methods, e.g., \cite{kumar2017online, khoshnami2018two, pillai2018mppt, yurtseven2021sensorless}, are also dependent on specific user-defined thresholds as they typically need a ``normal''  PV generation as the baseline to determine faults by comparison. Furthermore, the calculations required in the algorithms of these studies, e.g., \cite{kumar2017online, khoshnami2018two, yurtseven2021sensorless}, are usually computationally intensive. This means while these methods can be a practical solution for solar PV farms, implementing them for all residential rooftop PV systems demands massive processing power. Lastly, many of these methods rely on high-resolution data to detect potential issues in the systems \cite{kumar2017online, khoshnami2018two, pillai2018mppt, fadhel2019pv}, but such data may not be recorded for residential rooftop PV systems.

Addressing these issues, we propose an efficient and practical method for detecting malfunctions or abnormalities in rooftop PV systems in a geographically wide area based on a big data approach. The only input to our model is historical generation data of rooftop PV systems in a given area, e.g., a specific postcode, or a group of postcodes, during a few months, which are already available through smart metering infrastructure. As a result, no new sensors or smart devices are needed in this approach. This is thanks to the power of big data, which enables us to find anomalies by comparing the generation patterns of PV systems in a region with each other. Furthermore, using a computationally efficient complexity measure, i.e., weighted permutation entropy (WPE) \cite{Fadlallah2013}, it only takes a few minutes to identify the anomalies on a standard PC; thus, not necessary to use a high-performance computing device. Note that the proposed method cannot be used for just a couple of PV systems as it works based on the big data approach and requires data from at least a few PV systems in an area. Also, this method is not suitable for online fault detection since it requires data of at least a few days to detect anomalies. However, its efficiency and simplicity for a large number of rooftop systems in a large area can advise O\&M crews to detect the faults in the systems that could not be detected otherwise by online protection methods. Resolving such malfunctions leads to higher efficiency and extended lifespan, which saves PV owners hundreds of dollars annually and gives an advantage to the PV system service providers.

\section{The Proposed Methodology}
In this section, we explain how to calculate WPE and describe the step-by-step implementation of the proposed anomaly detection method for rooftop PV systems.
\subsection{WPE Calculation}
\label{WPECalc}
In order to calculate WPE of a time series of length $N$, $\{x_t\}_{t=1,...,N}$, for embedding dimension $d$ and time delay $\tau$, we firstly divide the time series into $N - (d - 1)\cdot\tau$ embedding vectors (i.e., sequences of values) of length $d$ and time delay $\tau$, i.e., $ X^{d,\tau}_t = (x_t,\,x_{t+\tau},\,...,\,x_{t+(d-1)\cdot\tau}) $ for $t=1,...,N - (d - 1)\cdot\tau$. Then, we assign each vector to a single permutation, $\pi_k$, in the set of possible permutations, $\Pi$, which includes all possible unique orderings of $d$ real numbers (so, there are $d!$ unique permutations in $\Pi$). In other words, we associate each sequence of values, $X^{d,\tau}_t$, to one permutation, $\pi_k$, based on the sequence's ordinal pattern, $ \pi_k \sim \phi(X^{d,\tau}_t)$. For example, if $X^{d,\tau}_t$ is \{4,3,7\}, then $\phi(X^{d,\tau}_t)$, the ordinal pattern of this sequence, is 2-1-3. 

To calculate the WPE value, we determine the weight of each vector as follows \cite{Fadlallah2013}:
\begin{gather}
w_t  =  \frac{1}{d} \sum\nolimits_{s=1}^{d}(x_{t+(s-1)\tau}-\bar{X}_{t}^{d,\tau})^2, \label{weight}
\end{gather} 
where $\bar{X}_{t}^{d,\tau}$ is the arithmetic mean of the values in the corresponding vector. 

For each $\pi_k \in \Pi$, the weighted probability of permutation $\pi_k$ occurring in time series $\{x_t\}$ is
\begin{gather}
P_w(\pi_k)  =  \frac{\sum_{t=1}^{N - (d - 1)\tau}\big(w_t.\psi(\phi(X_t^{d,\tau}),\pi_k)\big)}{\sum_{t=1}^{N - (d - 1)\tau}w_t}, \label{WPEprob}
\end{gather}
where $\psi(a,b)$ is 1 when $a=b$, and 0 otherwise. The numerator of this equation is the sum of all weights of the vectors that have the same pattern as permutation $\pi_k$. The denominator of this equation is the sum of the weights of all vectors in the time series. Finally, the weighted probability of all permutations is used to calculate WPE.

Using the above definitions, WPE for $d \geq 2$ is defined as
\begin{gather}
H_w^{d,\tau}(\{x_t\}) = - \sum\nolimits_{\pi_k \in \Pi }~P_w(\pi_k)~log_2 (P_w(\pi_k)) ~ . \label{WPE}
\end{gather}  

We normalise WPE by $log_2 (m!)$, and, hereafter, the normalised WPE is referred to as the ``WPE''. Authors in \cite{Bandt2002} recommended that, for practical purposes, the embedding dimension should be a number between 3 and 7 ($d\in\{3,4,5,6,7\}$). Additionally, to allow all possible patterns in the time series to appear in the analysis, we should select the dimension such that $d! << N $. For a finite time series, this allows an accurate estimation of the relative frequency of permutations. Thus, to determine the exact values of the frequencies, we must have $N \to + \infty$ \cite{Zunino2010}. However, considering the limitations of real-world time series and based on studies in various disciplines, authors in \cite{Riedl2013} recommended that the maximum dimension should be selected such that the length of the time series satisfies $N > 5d!$.

\subsection{Anomaly Detection}%
As WPE measures the complexity of a signal and is determined based on its existing patterns \cite{Bandt2002}, we expect that the generation of the rooftop PV systems in an area will have almost similar WPE profiles over time. This is because the PV generation significantly relies on solar irradiance and temperature, which vary similarly in a region as big as a neighbourhood or several postcodes. This is validated in the next section using real-world PV generation data. 

Given that WPE is an intrinsic feature of PV generation in a region, a PV system that exhibits significantly different WPE evolution over time can be detected as anomalous. Then, the O\&M team is alerted to inspect the anomalous system and resolve the issue. This would be useful for the O\&M team to easily pinpoint the systems that require an inspection and ensure the high efficiency and safety of the rooftop systems.

\begin{figure}%
	\centering
	\includegraphics[width=0.75\linewidth]{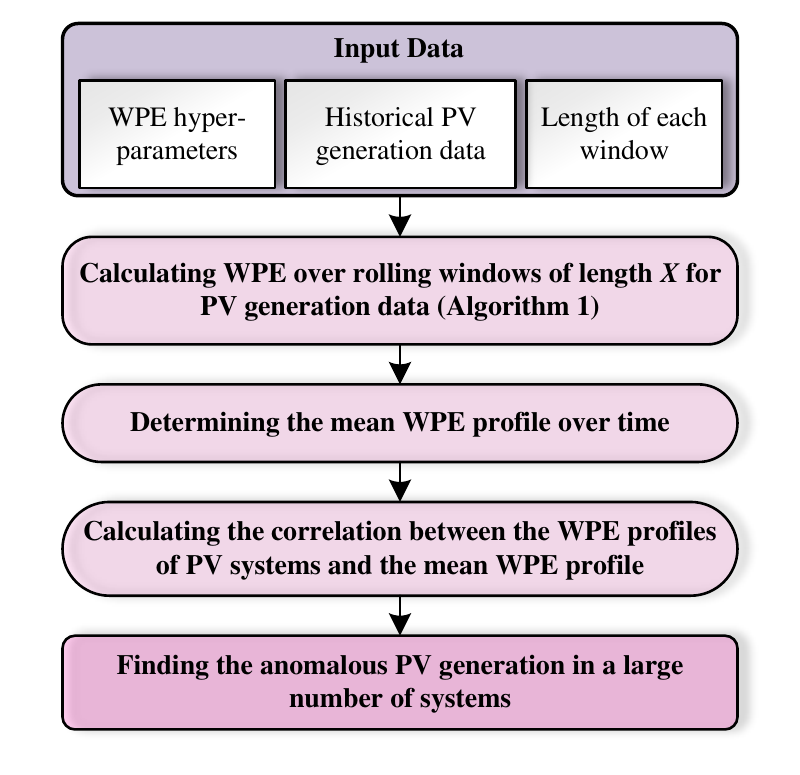}
	\caption{Flowchart of the proposed anomaly detection method.}
	\label{Flowchart}
\end{figure}

Our proposed method can be implemented step-by-step, as illustrated in Fig.~\ref{Flowchart}. The first step is to calculate the WPE of each PV generation time series over rolling windows of width $X$ (the width of each window should be large enough to satisfy the conditions discussed in Subsection II.A). These calculations can be performed by following the steps in Algorithm \ref{alg:1}, which computes WPE profiles of PV systems over time. After this step, the mean WPE profile of all PV systems in the given area is calculated. We then determine the correlation between each system's WPE profile and the mean WPE profile. The PV systems with a ``low'' correlation compared to others are outliers and need to be investigated further. This will be done by comparing the WPE profiles of these PV systems to the mean WPE profile to find the period when the anomaly occurred and then by analysing the PV generation during that period to understand the issue. It is worth mentioning that a suitable value for the low correlation threshold is 0.8 since the PV generation with a WPE profile of the correlation above this value has a highly similar generation pattern compared with the majority in the same area. Nevertheless, if there are a large number of PV systems available in a given region, a better approach is to determine the outliers, e.g, a correlation value is an outlier if it is more than the interquartile range of the correlation values below their first quartile. 
\begin{algorithm}
\caption{Calculating WPE over rolling windows}\label{alg:1}
Set the window width equal to $X$;

\For{each PV generation time series}{

    \Repeat{reaching the end of the time series}{
    
        Set the window starting and ending point;\\
        Calculate the WPE of the window time series as explained in Subsection II.A;\\
        Roll the window one step forward;
    }
}
\end{algorithm}

\section{Numerical Results}
\subsection{PV Generation Dataset}
In this study, we use a PV generation database of 1000 houses with rooftop PV systems across Australia \cite{solardata}, located in different states and various postcodes. The length of each time series is one year (from 1 January 2019 to 1 January 2020), and the sampling interval is five minutes. 

To prepare the dataset for analysis, the time series with more than 200 missing points have been removed. Time series with less than 200 missing samples were treated with the last observation carried forward method and used in the simulation studies. Also, we removed the time series that included significant negative or high generation curtailment for more than 7 months. In the end, we had 335 PV generation time series available for analysis. The following results are obtained from implementing WPE and the proposed method using Python and SQL. 

\subsection{Tuning the WPE Hyperparameters}
\label{ParameterSelect} 
\begin{figure}
	\centering
	\includegraphics[width=0.98\linewidth]{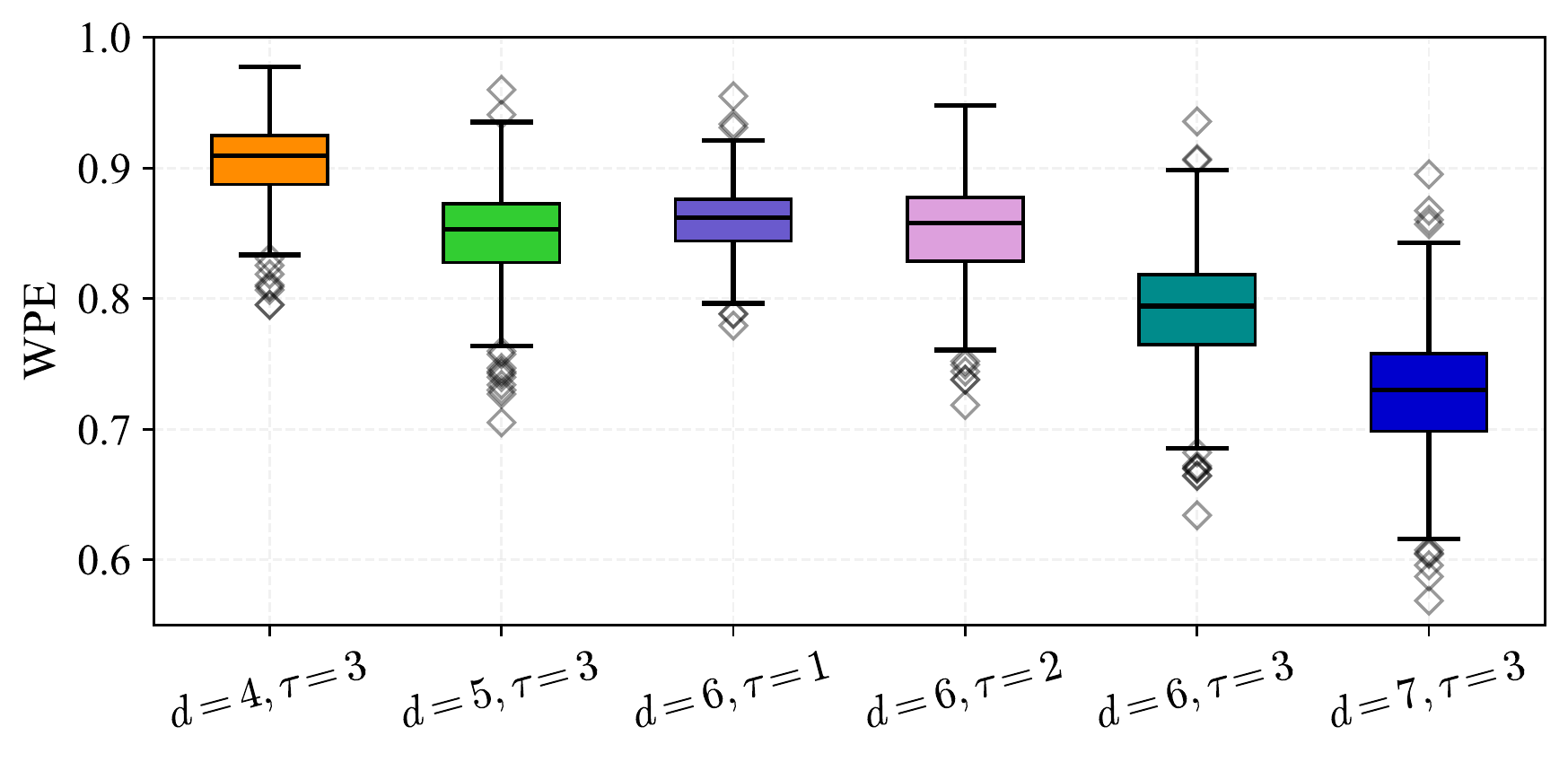}
	\caption{The boxplot representing the distribution of WPE values for different parameter settings.}
	\label{BoxPlot}
\end{figure}%
WPE calculation requires specifying the embedding dimension, $d$, and the time delay, $\tau$. Tuning these parameters is particularly important for two reasons: 1) to capture all the important patterns in the time series and 2) to provide a good estimation of permutation frequencies. The former can be achieved when the embedding dimension, $d$, is large enough to discover all admissible patterns in the time series. This is because a bigger dimension means more permutations, to which we can assign the time series sequences. On the other hand, the dimension should not be too large to ensure that we can gather reasonable statistics over different permutations, providing a good estimation of permutation frequencies ($P_w(\pi_k)$).

Selecting an appropriate time delay is more arbitrary compared to that of the embedding dimension. In the literature, it is usually set to an arbitrary value without any justification. However, selecting a time delay higher than one essentially depends on our knowledge of the underlying system dynamics. In our dataset, the sampling interval is five minutes. Therefore, the time series resolution is high enough to set the time delay to one, two, or three. However, choosing a greater time delay means we cannot discover repetitive sub-patterns in the high-resolution data, i.e., missing high-resolution dynamics. Also, selecting a time delay equal to one means that the length of the embedding vectors is 30 minutes at best (for the case with a dimension equal to 7), which might be too short to capture repetitive structures in PV generation data. Therefore, the best option for the time delay in our study is two and three.

Accordingly, to find the best hyperparameters in this application, we have done a sensitivity analysis on different sets of WPE hyperparameters, the results of which are reported in Fig. \ref{BoxPlot}. It shows a box-and-whisker plot for the parameter settings that can provide the most suitable WPE values for our dataset. In this figure, the solid line represents the median, the boxes enclose the first and the third quartiles of WPE distributions, and the whiskers cover the entire WPE distribution, excluding the outliers. We can observe that as the time delay increases from 1 to 3, the range of WPE values within the box and whiskers in the figure increases, which means more distinguishability among WPE values of the time series. This observation and our previous argument about the suitability of a time delay equal to two or three led us to set the time delay to three in the rest of the simulation studies.

According to Fig. \ref{BoxPlot}, similar to the time delay, higher embedding dimensions lead to wider distributions for WPE values, which improves the ability to distinguish PV systems based on WPE values. However, as the dimension increases, the difference between the interquartile ranges decreases such that the range of WPE values (excluding the outliers) for $d=7$ is slightly longer than that of $d=6$. It means that using a higher dimension ($d=7$) in the WPE calculation does not improve the distinguishability of different time series while increasing the computational time by almost four times on average. Also, when the embedding dimension is set to 7, there are 7 times more possible permutations. Because the time series has a relatively short length, the estimation accuracy for each permutation's frequency would be lower than the case with $d = 6$. Thus, we use $d = 6$ in our simulation studies.

\subsection{Comparing PV Generation in Different Postcodes}
\begin{figure}
	\centering
	\includegraphics[width=0.7\linewidth]{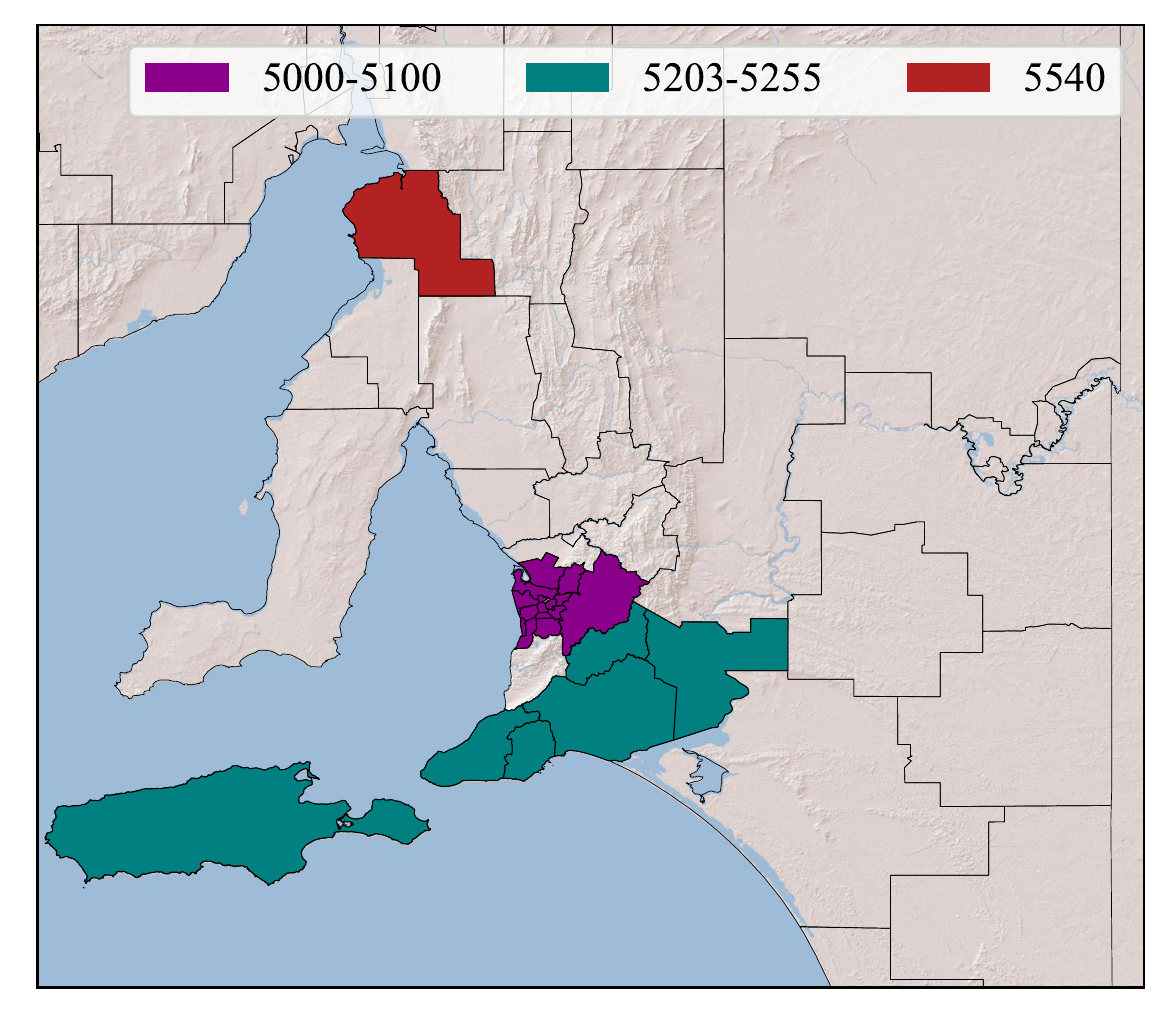}
	\caption{The SA regions where the rooftop PV systems are installed.}
	\label{SAmap}
\end{figure}%
\begin{figure}[t]%
    \captionsetup[subfigure]{aboveskip=-1pt,belowskip=-1pt}
    \centering
    \begin{subfigure}{\linewidth}%
	\includegraphics[width=0.94\linewidth]{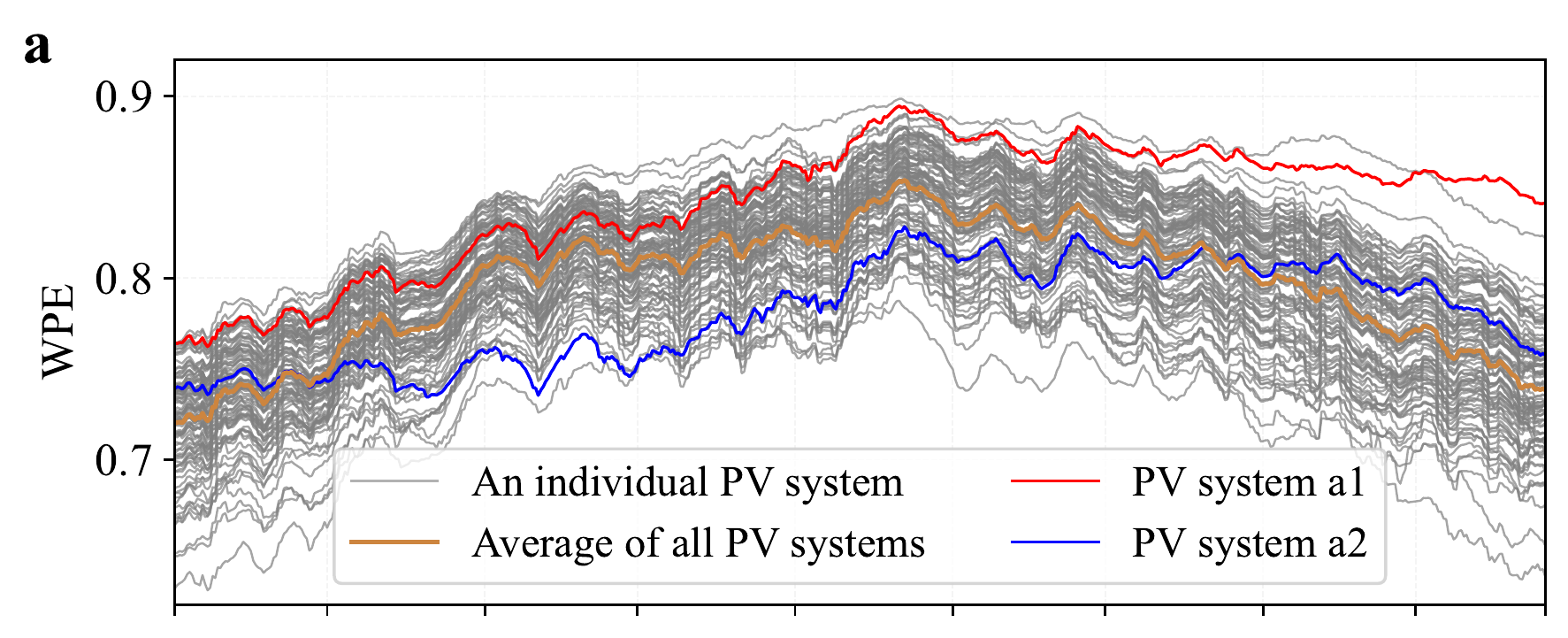}%
	{\phantomsubcaption\label{fig:WPEgraphsofSA_a}}%
    \end{subfigure}\\ \vspace{-0.1cm}
    \begin{subfigure}{\linewidth}%
	\includegraphics[width=0.94\linewidth]{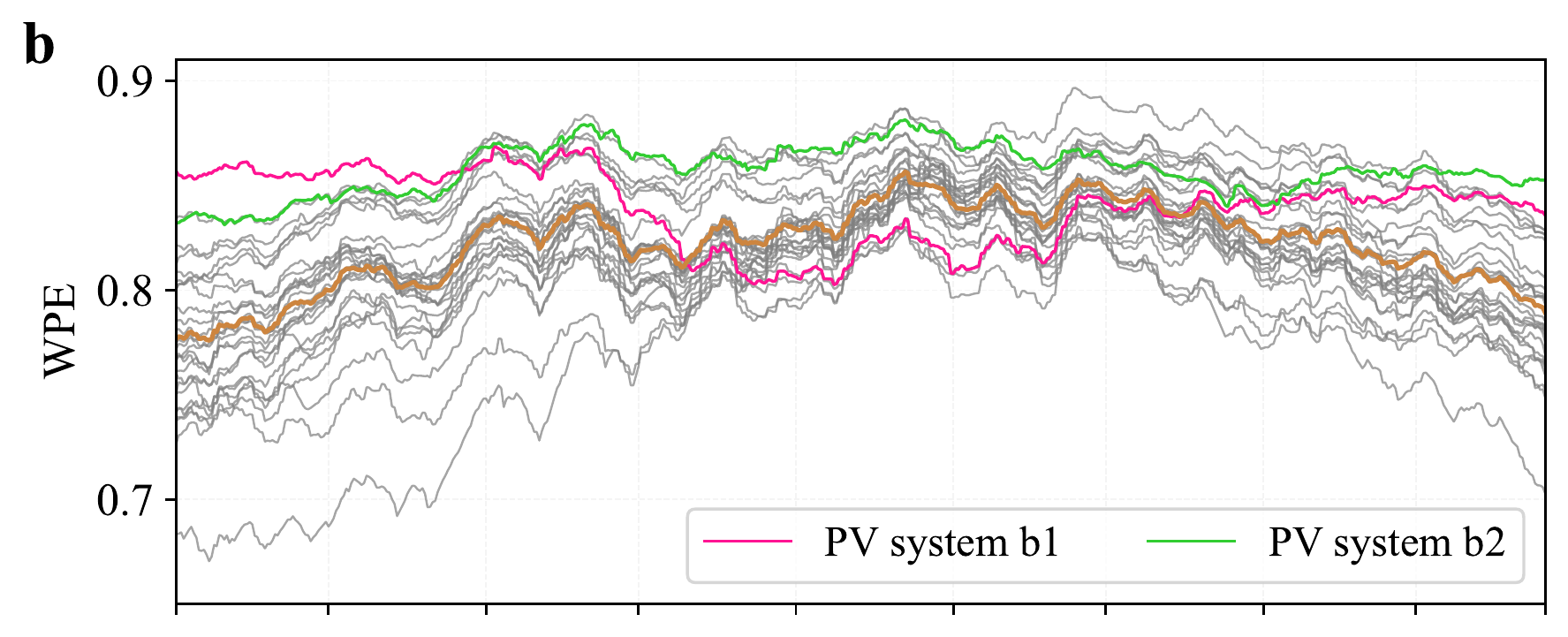}%
	{\phantomsubcaption\label{fig:WPEgraphsofSA_b}}%
    \end{subfigure}\\ \vspace{-0.1cm}
    \begin{subfigure}{\linewidth}%
	\includegraphics[width=0.94\linewidth]{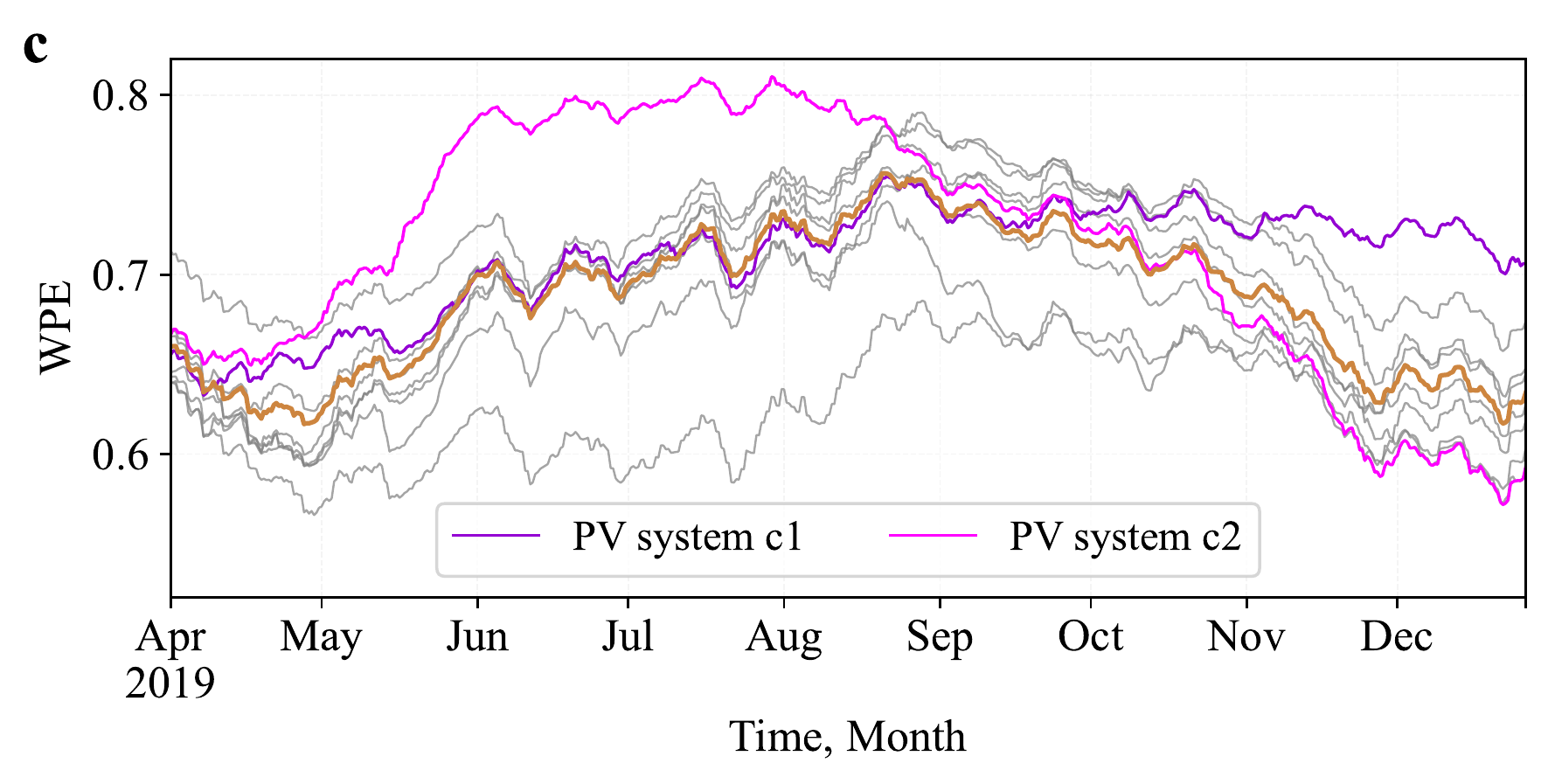}%
	{\phantomsubcaption\label{fig:WPEgraphsofSA_c}}%
    \end{subfigure}%
    \caption{WPE profiles of a three-month rolling horizon over a year for the generation of rooftop PV systems located in the postcode(s) of: (\textbf{a}) 5000--5100, (\textbf{b}) 5203--5255, and (\textbf{c}) 5540.}
    \label{WPEgraphsofSA}
\end{figure}
\begin{figure}[t]
	\centering
	\includegraphics[width=0.9\linewidth]{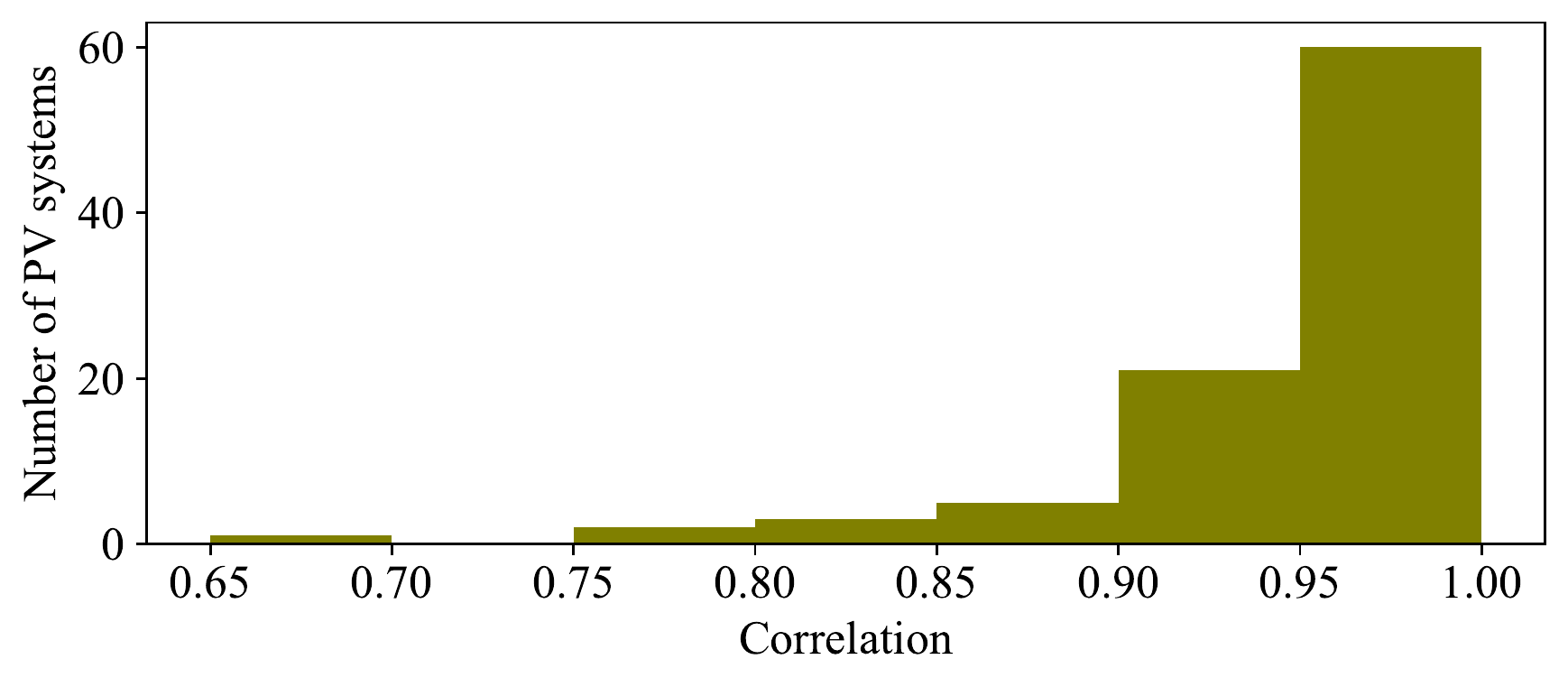}
	\caption{The histogram of correlations between the generation WPE of all PV systems and the mean WPE.}
	\label{Hist}
\end{figure}%
By leveraging the similarity of WPE profiles from PV systems across a region, we hope to identify anomalous systems. To this end, we created three groups of rooftop PV systems based on the regions in which they are installed. The three groups are postcodes 5000-5100, 5203-5255, and 5540. These three regions are shown in Fig. \ref{SAmap}.
\begin{figure*}[t]%
    \captionsetup[subfigure]{aboveskip=-1pt,belowskip=-1pt}
    \centering
    \begin{subfigure}{0.76\columnwidth}%
	\includegraphics[width=1\linewidth]{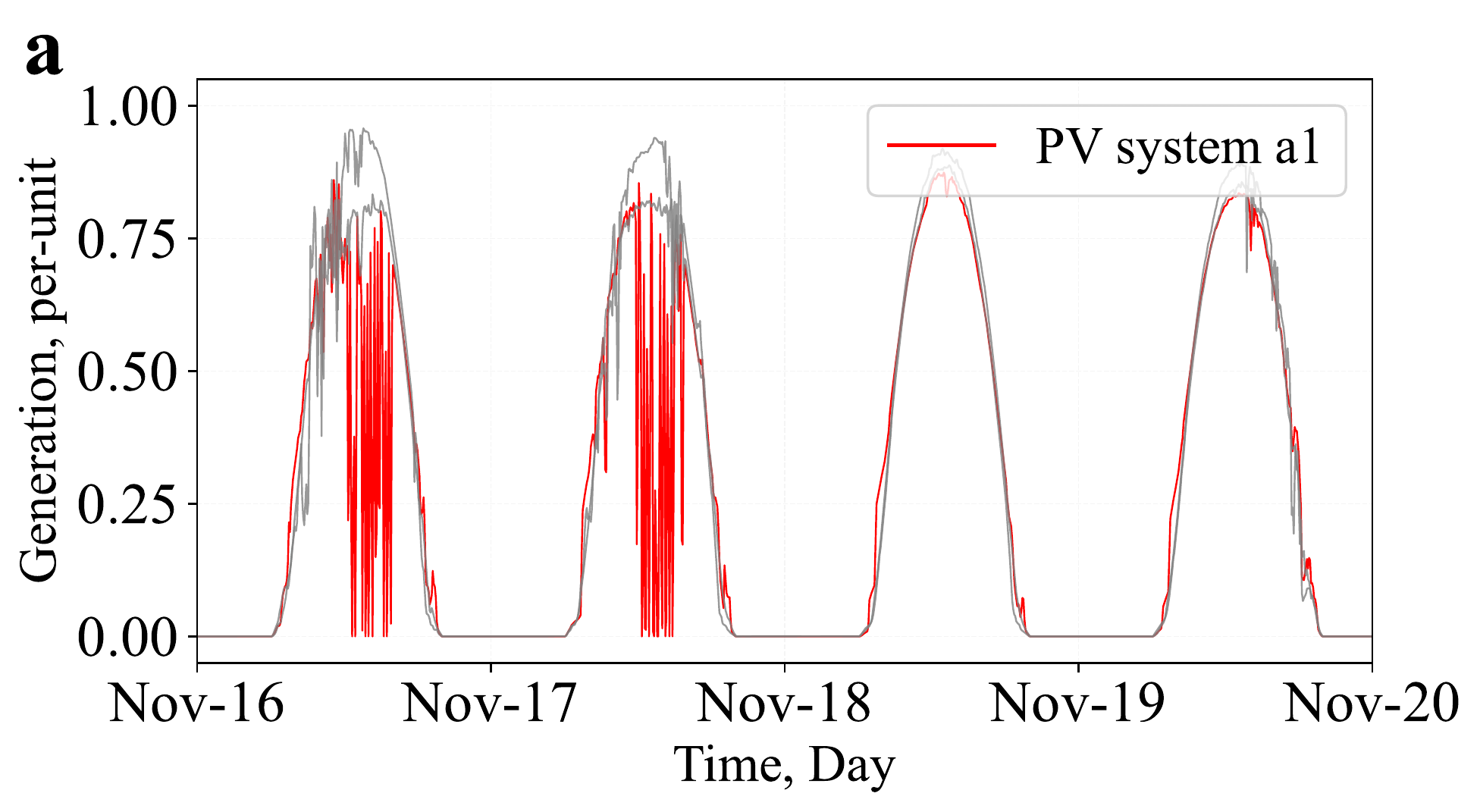}
	{\phantomsubcaption\label{fig:3Gen_a}}
    \end{subfigure}%
    \begin{subfigure}{0.76\columnwidth}%
	\includegraphics[width=1\linewidth]{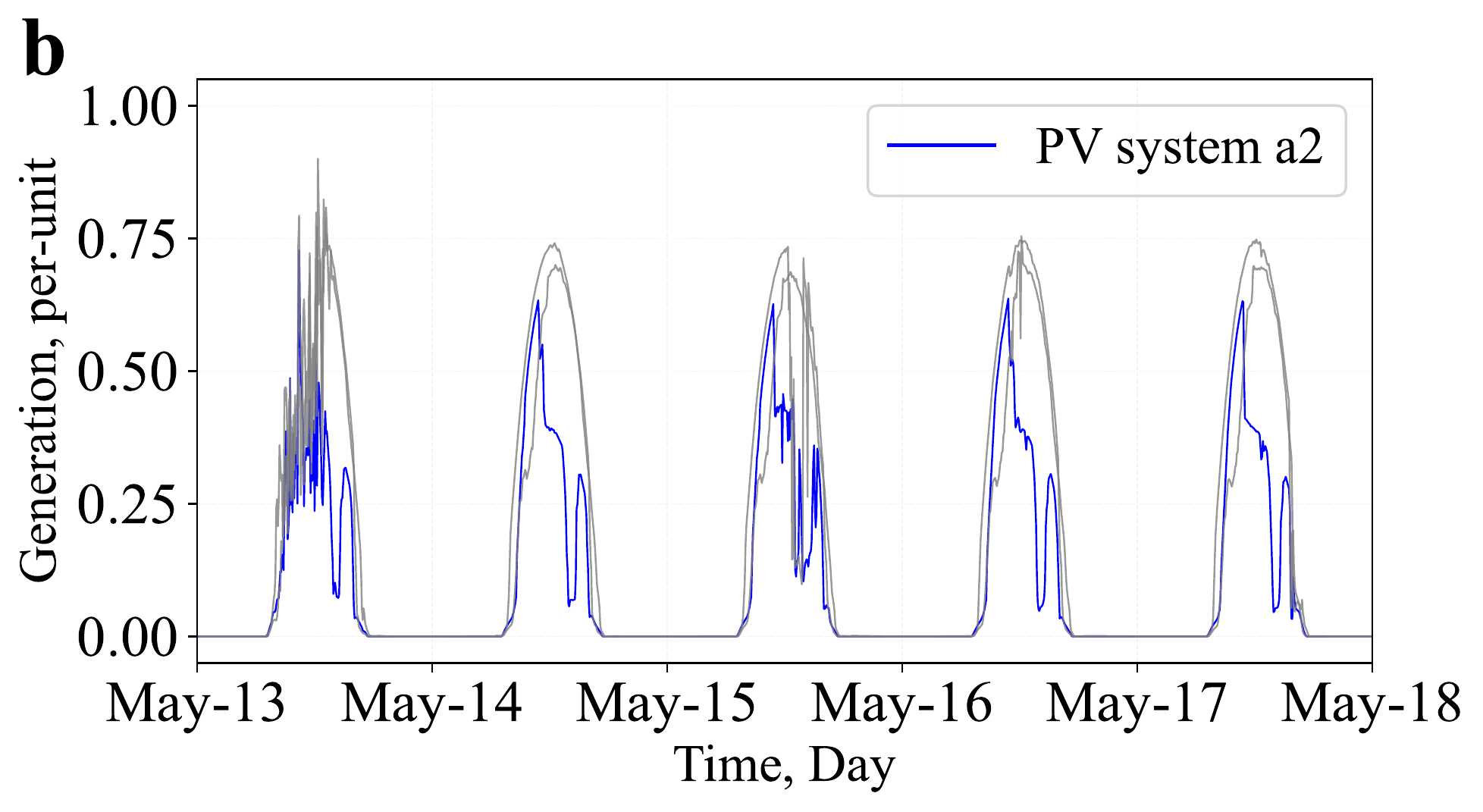}
	{\phantomsubcaption\label{fig:3Gen_b}}
    \end{subfigure}\\
    \begin{subfigure}{0.76\columnwidth}%
	\includegraphics[width=1\linewidth]{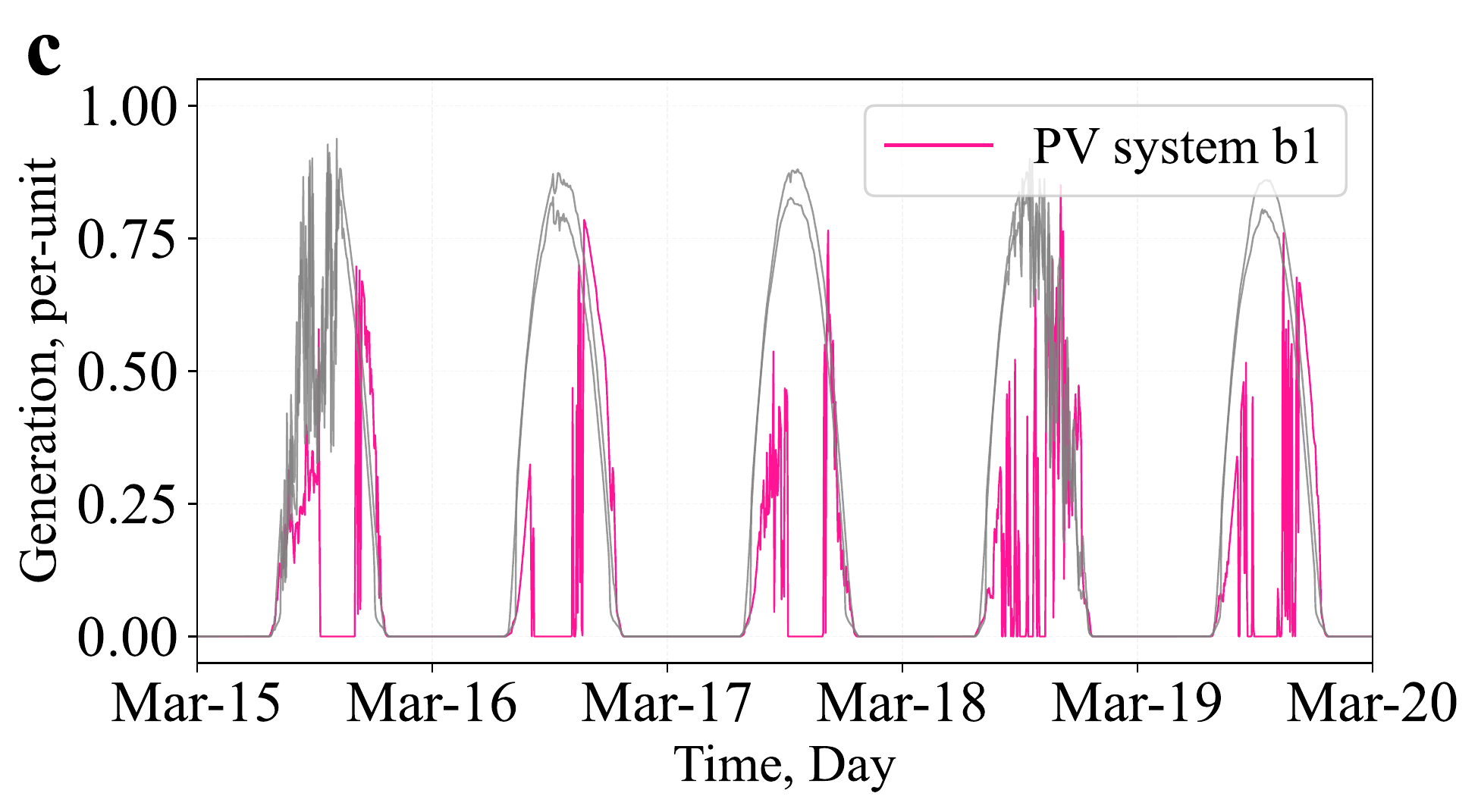}%
	{\phantomsubcaption\label{fig:3Gen_c}}
    \end{subfigure}%
    \begin{subfigure}{0.76\columnwidth}%
	\includegraphics[width=1\linewidth]{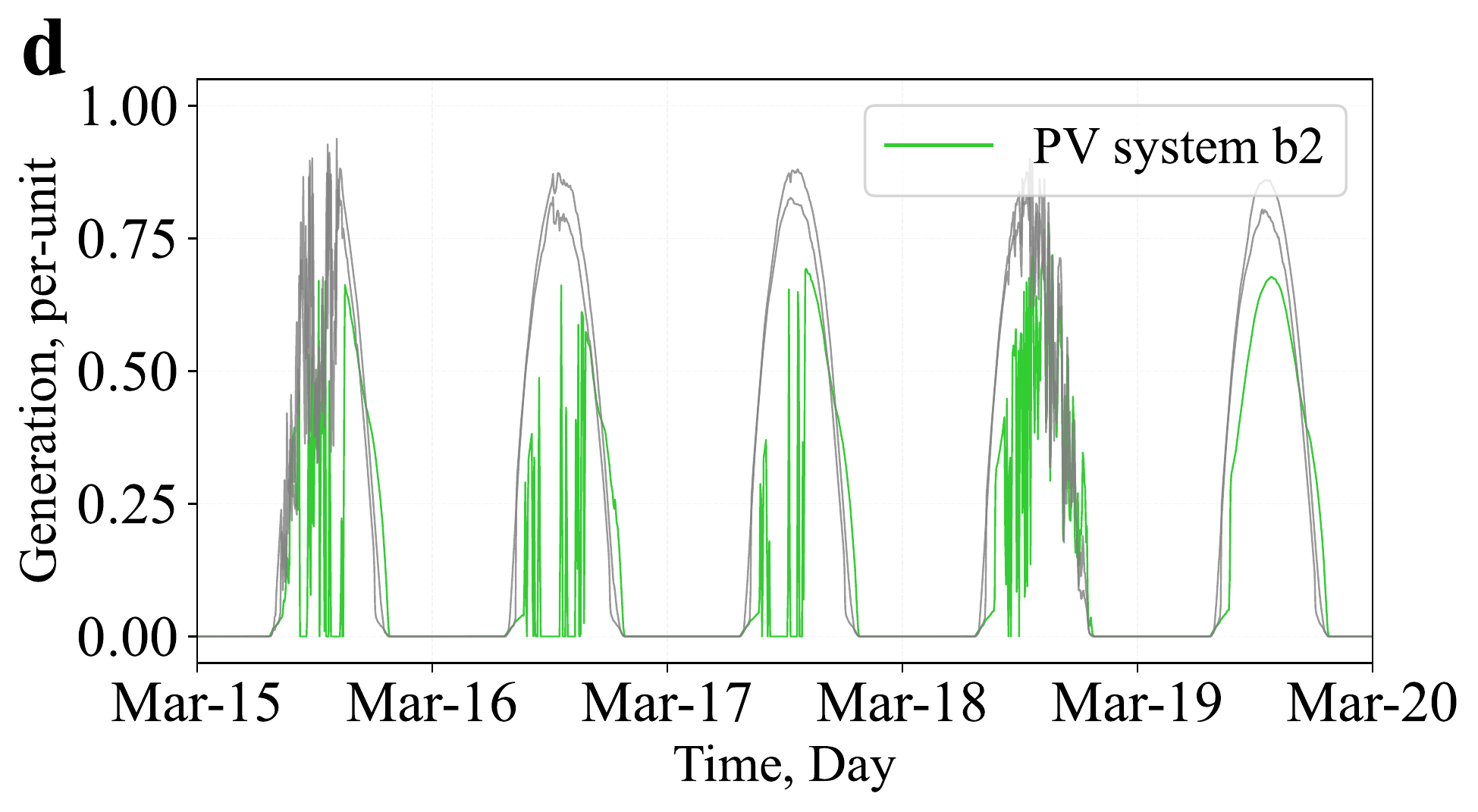}%
	{\phantomsubcaption\label{fig:3Gen_d}}
    \end{subfigure}\\
    \begin{subfigure}{0.76\columnwidth}%
	\includegraphics[width=1\linewidth]{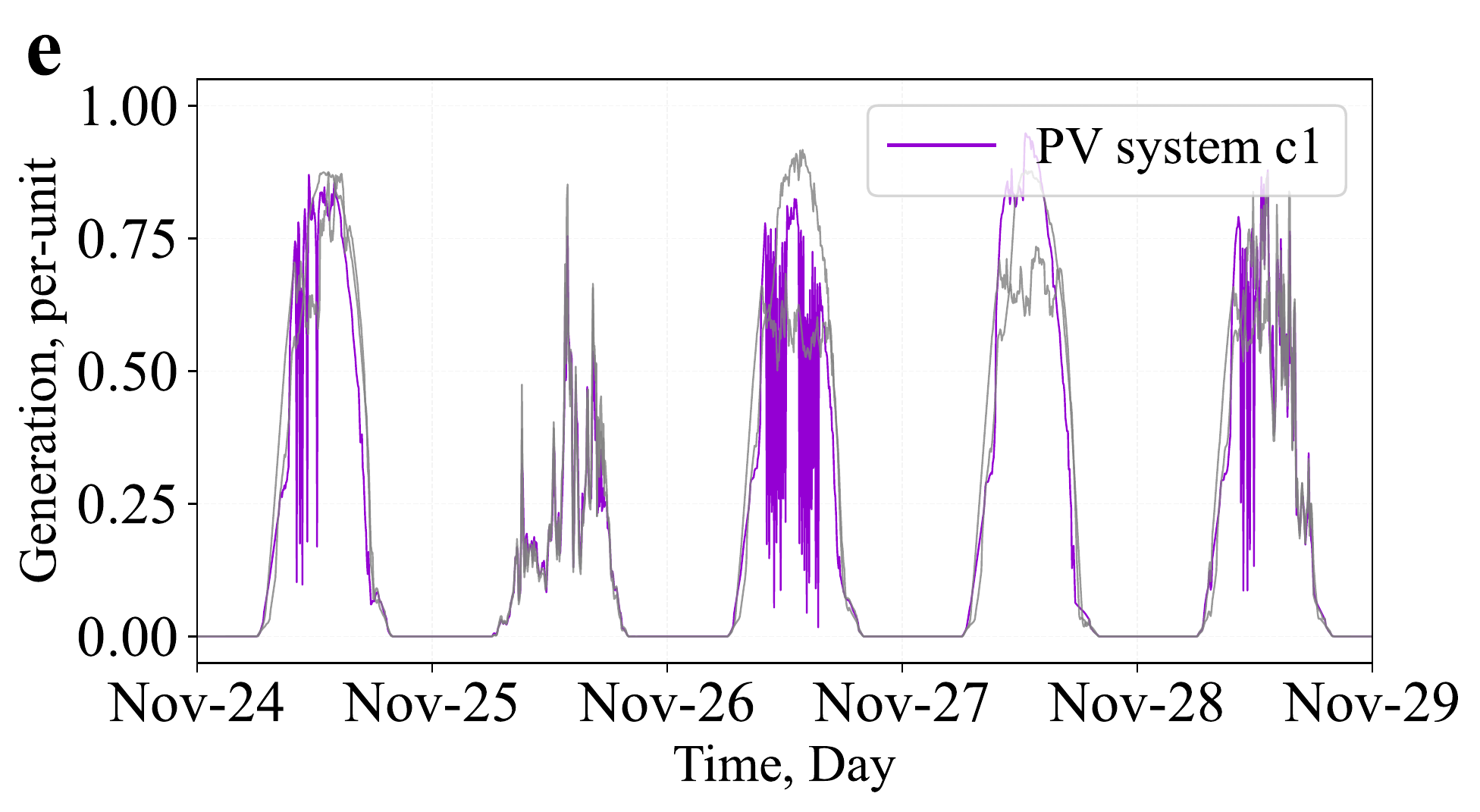}
	{\phantomsubcaption\label{fig:3Gen_e}}
    \end{subfigure}%
    \begin{subfigure}{0.76\columnwidth}%
	\includegraphics[width=1\linewidth]{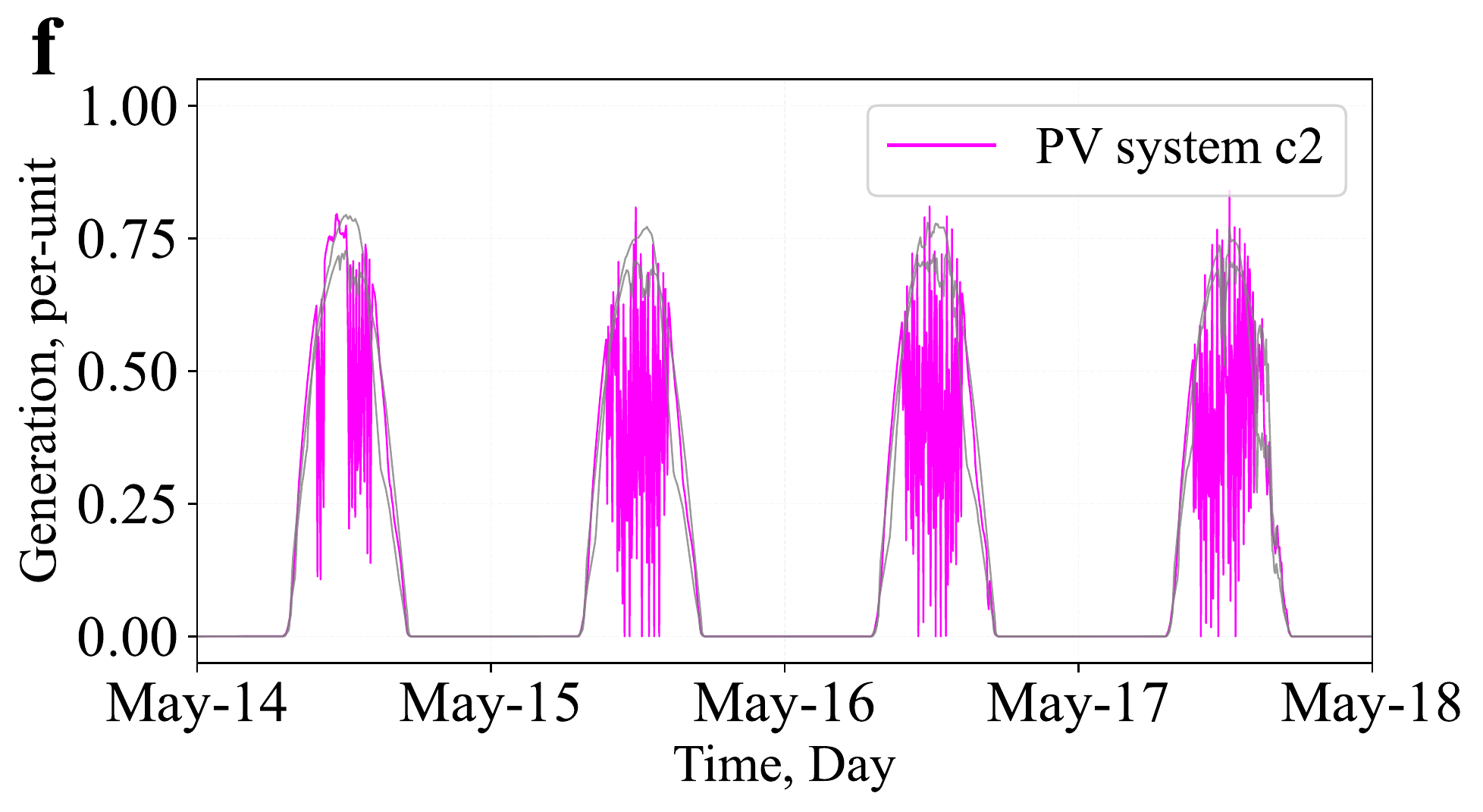}
	{\phantomsubcaption\label{fig:3Gen_f}}
    \end{subfigure}
    \caption{Detected anomalies using WPE profiles of PV generation in SA. The generation profile of PV systems (\textbf{a}) a1, (\textbf{b}) a2, (\textbf{c}) b1, (\textbf{d}) b2, (\textbf{e}) c1, and (\textbf{f}) c2 in comparison with the generation of two random PV systems at the same region. }%
    \label{3Gen}
    \vspace{-0.4cm}
\end{figure*}%

The WPE profiles of the PV generation over three-month rolling windows for the PV systems in these three regions are shown in Fig. \ref{WPEgraphsofSA}. We set the length of windows, $X$, to 3 months, considering the WPE hyperparameters determined in the previous subsection, as well as the WPE calculation constraints discussed in Subsection II.A. It is interesting to observe that the WPE of the PV generation of the houses that are located in close proximity display extremely similar trends over time. This indicates that the changes in the PV generation WPE over time significantly depend on the location of the PV systems, which is consistent with our expectation of WPE dependence on the existing patterns in a time series. In the figure, we can also observe that the WPE profiles of the systems in a larger region, like postcodes 5203-5255, are slightly less correlated compared to those in a smaller region, postcodes 5000-5100. These observations validate our hypothesis that WPE is an intrinsic feature of the PV generation time series that highly depends on its location. To show an example of this high correlation among WPE profiles in a region, Fig. \ref{Hist} depicts the histogram of correlations between the WPE profiles of system in postcodes 5000-5100 and the mean of all WPE profiles in the same region. While the WPE profiles of most PV systems are highly correlated with the mean WPE profile (the great majority have a correlation higher than 0.8), there are few systems with significantly lower correlations. Based on this observation, we can identify anomalous PV systems among a large number of them in a given region, which is investigated further in the next subsection.
\subsection{Finding Anomalous Rooftop PV Systems}
In this subsection, we use the similarity of WPE profiles of PV systems in a specific region to find anomalies. To show a few examples of this, we selected two PV systems in each region shown in Fig. \ref{WPEgraphsofSA}, whose WPE profiles had the least correlation with the mean WPE profile of that region (all 6 systems have correlations less than 0.8 with the mean WPE profile in their respective regions). These WPE profiles are shown in different colours. We can see that, in most cases, the WPE profiles were significantly different to the mean profile only during a few months of the year. In Fig.~\ref{fig:WPEgraphsofSA_a}, the WPE profiles of PV systems a1 and a2 show different behaviours from October to the end of the year and June to August, respectively. Also, we can see the different behaviours from PV systems b1 and b2 in Fig.~\ref{fig:WPEgraphsofSA_b} compared to other PV systems in that area for the most part of the year. Lastly, in Fig.~\ref{fig:WPEgraphsofSA_c}, we can observe that from May to August and October to the end of the year, the WPE profiles of PV systems c1 and c2 change differently compared to the mean WPE in that area. 

To demonstrate that these PV systems had significantly abnormal generation and our method could successfully detect them, we compared the generation of these PV systems, coloured in Fig. \ref{WPEgraphsofSA}, with two randomly-selected normal PV systems in the same region. For this purpose, we depicted the generation profiles of these PV systems in six subfigures for a few days in Fig.~\ref{3Gen}, during which the anomalies are identified, alongside two typical PV generation profiles from the same area. It is important to note that, in this figure, the base value for per-unit generation is the maximum value for the whole year. We can easily see the differences between the generation profiles of the anomalous PV systems and their deemed normal counterpart systems. In Fig.~\ref{fig:3Gen_b}, for example, we can see abnormal but repetitive patterns in the generation of PV system a2, which is consistent with its lower WPE (because the existence of more repetitive patterns means lower WPE) shown in Fig.~\ref{fig:WPEgraphsofSA_a}. While it looks like partial shading incidents, a closer inspection is required to identify the root cause. Also, in Figs.~\ref{fig:3Gen_c} and \ref{fig:3Gen_d}, we can see that the generation of PV systems b1 and b2 has been curtailed (whether intentionally by the network export limit or due to a malfunction in the inverter operation) in some parts of the days and is rapidly fluctuating in other parts of the same days. This is the reason for their higher WPE compared to typical PV systems in the same region. Lastly, Figs. \ref{fig:3Gen_e} and \ref{fig:3Gen_f} reveal that the generation of PV systems c1 and c2 on most days fluctuates rapidly, unlike other PV systems in the same area, leading to relatively higher WPE values during the time that these anomalies happened. 

These examples demonstrate that we can successfully detect even small-scale anomalies in a large number of PV systems using this method. While anomalies that lead to a significantly lower generation than the average or median of PV systems in an area can be found without using the proposed method, the ones we showed here could be difficult to detect by using such intuitive methods. This is because they are mainly related to the anomalies in the generation pattern rather than the amount. In addition, these anomalies have happened only in specific months while the system behaved normally in the rest of the year, meaning that the average generation of a PV system might not be significantly different even with the fault. For instance, while our method could easily find the anomalies in PV systems a1 and a2 among 92 systems located in postcodes 5000-5100, their average generation values (0.170 and 0.159) are in the interquartile range of all systems and are not much different to the mean (0.171) in that region. It is worth reminding that the PV generation expressed in per-unit with the maximum generation during the year as the base value. These results demonstrate that the proposed method can help rooftop PV service providers determine the systems experiencing undetected faults.

\section{Conclusion}
This paper presented an efficient, easy-to-use, and cheap anomaly detection method for rooftop PV systems using big data and permutation entropy. To do so, we used a time series complexity measure called WPE, which quantifies an intrinsic property of PV generation time series that depends on the system location. Using this feature, we could specify whether the generation patterns of a PV system are different to other systems in the same area over the same time period and find the anomalous systems. To validate this method, we applied it to a real-world rooftop PV generation dataset that included the generation of rooftop systems in different regions of SA. Our results demonstrated how this method could identify potential faults in PV systems installed in those regions. As this method does not require installing new sensors or devices and can be readily used for a large number of systems in a region thanks to its low computational overhead, it would be an affordable solution for PV service providers and their existing customers. While the method cannot find the root cause of faults, which should be done by a closer inspection, pinpointing potential faults and resolving them would mean higher efficiency of rooftop PV generation and a longer lifetime of the PV systems. This would avoid wasting clean solar energy and increase the average return on investment of installing rooftop PV systems.

\section*{Acknowledgement}
This research project is financially supported by the School of Electrical and Electrical Engineering at The University of Adelaide through a PhD scholarship.

\bibliographystyle{IEEEtran}
\bibliography{Paper.bib}

\end{document}